\pdfoutput=1

\documentclass[11pt]{article}

\usepackage{emnlp2021}

\usepackage{times}
\usepackage{latexsym}
\usepackage{booktabs, multirow}
\usepackage[export]{adjustbox}
\usepackage{enumitem}

\usepackage[T1]{fontenc}

\usepackage[utf8]{inputenc}

\usepackage{microtype}

\title{Numerical reasoning in machine reading comprehension tasks: are we there yet?}

\author{Hadeel Al-Negheimish$^1$, 
 Pranava Madhyastha$^{2,1}$ \and Alessandra Russo$^1$ \\
  $^1$ Imperial College London \quad $^2$ City, University of London \\
  \texttt{\{halnegheimish,a.russo\}@imperial.ac.uk}, \\  \texttt{pranava.madhyastha@city.ac.uk}
 }

\begin{document}
\maketitle
\begin{abstract}

Numerical reasoning based machine reading comprehension is a task that involves reading comprehension along with using arithmetic operations such as addition, subtraction, sorting, and counting. The DROP benchmark~\citep{Dua2019DROP} is a recent dataset that has inspired the design of NLP models aimed at solving this task. The current standings of these models in the DROP leaderboard, over standard metrics, suggest that the models have achieved near-human performance. However, does this mean that these models have learned to reason? In this paper, we present a controlled study on some of the top-performing model architectures for the task of numerical reasoning. Our observations suggest that the standard metrics are incapable of measuring progress towards such tasks. 

\end{abstract}

\section{Introduction}
\label{sec:intro}

Machine reading comprehension (MRC) primarily involves building automated models that are capable of answering arbitrary natural language questions on a given textual context such as a paragraph. 
Solving this problem should in principle require a fine-grained understanding of the question and comprehension of the textual context to arrive at the correct answer. 
Designing an MRC benchmark is challenging, as it is easy to inadvertently craft questions that allow models to exploit cues that allow them to bypass the intended reasoning required \citep{gardner-etal-2019-making}.

Recent advances in NLP, currently dominated by transformer-based pre-trained models, have resulted in models that indicate, when measured with standard metrics, human-like performance on a variety of benchmarks over leaderboards for MRC\footnote{for eg., https://leaderboard.allenai.org/}. 
In this paper, we focus on numerical reasoning based MRC and investigate DROP \citep{Dua2019DROP}, a recent benchmark designed to measure complex multi-hop and discrete reasoning, including numerical reasoning\footnote{We refer the reader to \citet{thawani-etal-2021-representing}, for a recent survey on numeracy in NLP.}. In contrast to single-span extraction tasks, DROP allows sets of spans, numbers, and dates as possible answers. We are particularly interested in numerical questions, for eg., `How many years did it take for the population to decrease to about 1100 from 10000?' which requires extracting the corresponding years for the associated populations from the given passage, followed by computing the time difference in years. The benchmark has inspired the design of specialized BERT and embedding-based NLP models aimed at solving this task, seemingly achieving near-human performance (evaluated using F1 scores) as reported in the DROP leaderboard.

In this work, we investigate some of DROP's top-performing models on the leaderboard in order to understand the extent to which these models are capable of performing numerical reasoning, in contrast to relying on spurious cues. 
We probe the models with a variety of perturbation techniques to assess how well models \emph{understand} the question, and to what extent such models are basing the answers on the textual evidence. 
We show that the top-performing models can accurately answer 
a significant portion (with performance exceeding 35\%---61\% F1) of the samples
even with completely garbled questions. We further observe that, for a large portion of comparison style questions, these models are able to accurately answer without even having access to the relevant textual context. These observations call into question the evaluation paradigm that only uses standard quantitative measures such as F1 scores and accuracy. The ranking on the leaderboards 
can lead to a false belief that NLP models have achieved human parity in such complex tasks. We advocate the community to move towards more comprehensive analyses especially for leaderboards and for measuring progress.

\section{Dataset and Models}
In this section, we briefly highlight the dataset and models under consideration.
\label{sec:dataset}
\begin{table}[t]
\centering
\adjustbox{width=\columnwidth}{%
\begin{tabular}{@{}lllllll@{}}
\toprule
        & \multicolumn{2}{c}{\textbf{Dev} (9,536)} & \multicolumn{2}{c}{\textbf{Num}  (6,848)} &
        \multicolumn{2}{c}{\textbf{Test}  (9,622)}\\ \midrule
\textbf{Model}   & EM              & F1             & EM              & F1             & EM              & F1   \\ \midrule
BOW-Linear & 8.61           & 9.35          & 10.53           &10.77     &   & \\\midrule
NAQANet & 41.72           & 45.70          & 44.50           & 45.65       &44.24    &47.24   \\
MTMSN   & 70.54           & 74.95          & 77.37           & 78.01       &75.88    &79.99   \\
NeRd    & 73.86           & 77.24          & 78.49           & 79.02       &78.33    &81.71    \\
GenBERT & 63.18           & 67.60          & 72.55           & 73.25       &68.60    &72.35   \\
TASE    & 73.82           & 77.96          & 78.62           & 79.22       &80.42    &83.62    \\
NumNet+ & 73.56           & 77.70          & 79.15           & 79.80       &81.52    &84.84   \\\midrule
\textbf{Human}  &          &           &           &        &94.09 & 96.42\\
\bottomrule
\end{tabular}
}
\caption{Performance on a) devset (\textbf{Dev}); b) \texttt{numset} (\textbf{Num});  c) hidden testset (\textbf{Test}) (from leaderboard).}
\label{table:baseline}
\end{table}

\paragraph{Dataset}
DROP \citep{Dua2019DROP} includes questions with various types of reasoning, of which we are interested here in numerical reasoning. 
We filter the provided \emph{dev} set to only include questions that require numerical reasoning; this was done by first including all answers with type \emph{number}, and then augmenting with comparison questions, filtered heuristically based on whether it contains a comparative adjective or a comparative adverb. 5850 questions were \emph{number} and 998 were \emph{comparison}. We call this \texttt{numset} and use it in this paper as a basis for all experiments\footnote{all variations will be publicly released for reproducibility}. 
\paragraph{Models}
We include all publicly available models that appear in the DROP leaderboard\footnote{https://leaderboard.allenai.org/drop/submissions/public}, this includes NAQANet \citep{Dua2019DROP}, MTMSN \citep{hu-etal-2019-multi}, NeRd \citep{Chen2020Neural}, GenBERT \citep{geva-etal-2020-injecting}, TASE \citep{segal-etal-2020-simple} and NumNet+\citep{ran-etal-2019-numnet}, in addition to a simple logistic regression bag-of-words model to ground our results. With the exception of NumNet+, which has been trained on our local machines from a published codebase, we use the provided model checkpoint by the corresponding authors. All of the included models are based on the transformer architecture \citep{Vaswani-Attn}. 
They vary, however, on how they tackle the task: NeRd solves the problem by generating a program from a domain-specific language; GenBERT augments the language model pre-training procedure by adding two more stages, pre-training with numerical data and pre-training with numeric textual data; the rest rely on specialized modules to solve each of the different question types. The counting module frames the task as a multi-class classification problem of numbers 0-9, whilst the arithmetic module assigns a zero or a sign to each number in the passage and sums it up. Finally, they also differ in the encoder, where NAQANet is based on GloVe \citep{pennington2014glove} embeddings; MTMSN, NeRd and GenBERT use BERT-uncased \citep{devlin2018bert} (\textit{Large} variation for the first two, whereas the last is only available as \textit{Base}); and TASE and NumNet+ use RoBERTa$_{\small {\textsc{Large}}}$\citep{liu2019roberta}. 
We note that, while some of the models obtain human parity F1-scores, these models are not public. However, with their corresponding descriptions, these models are markedly similar to the models evaluated in this paper and we believe that our observations hold on these models too. 

Table~\ref{table:baseline} shows the performance of the models on the dev set, \texttt{numset}, and the test set scores as reported in the leaderboard. 
Note that scores for the first two columns and the ones reported in the rest of the paper are based on considering only the main annotator's answer as gold, whereas the official evaluation script considers all annotations. 
This is done to clearly track changes in output after input perturbations.

\section{Evaluating question understanding}
\label{sec:question-understanding}
Evaluating if a model \emph{understands} questions is a non-trivial task. In this paper, we probe the performance of the models in the following two ways: evaluation of the models with question permutation and investigating the affinity to question class. 

\paragraph{Question permutation experiment}
\begin{figure}[hb]
\centering
\adjustbox{width=\columnwidth}{%
\includegraphics[width=\columnwidth]{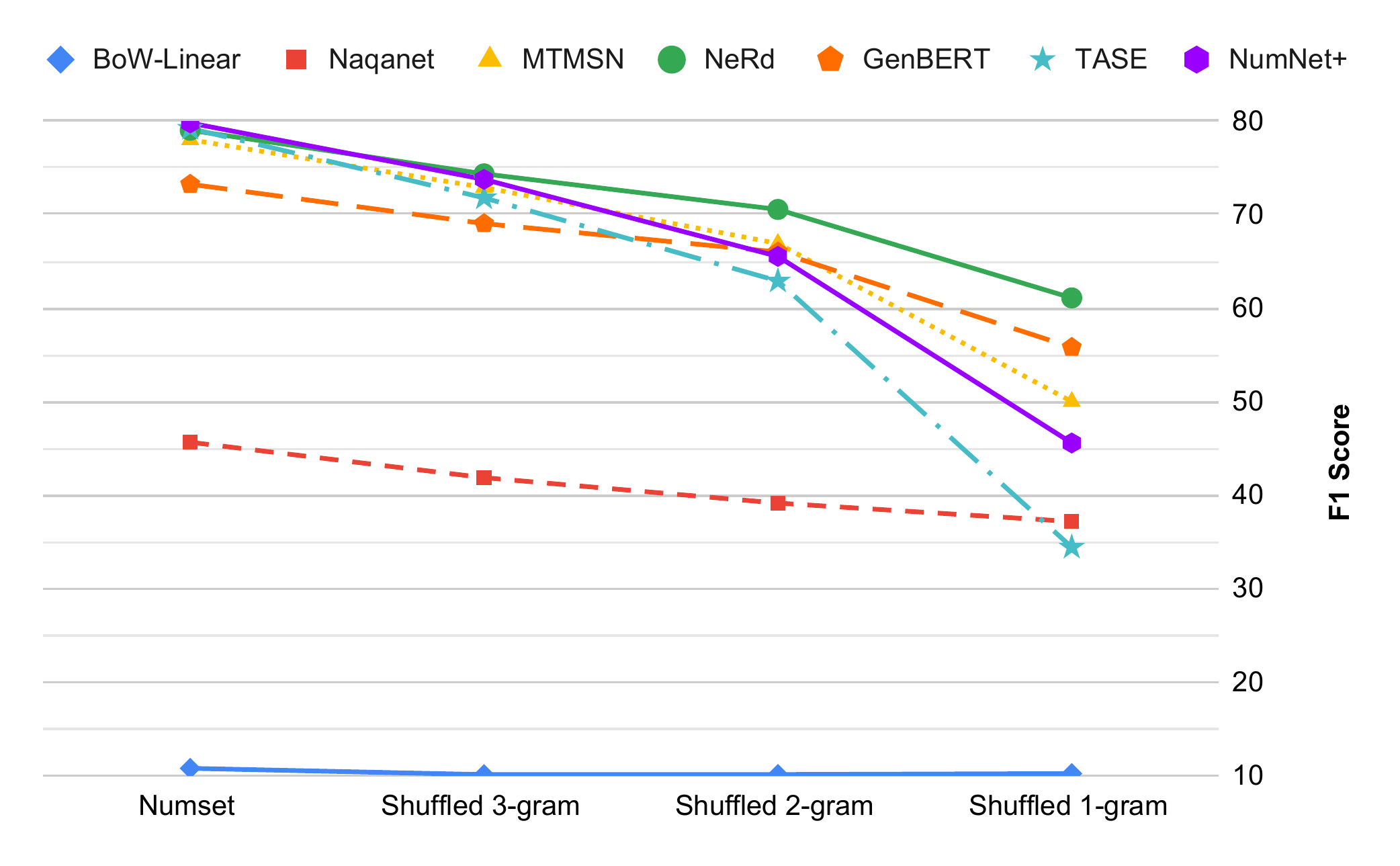}
}
\caption{F1 scores on \texttt{numset} and \{3,2,1\}-gram shuffles. We note that the results are stable over 5 different random permutations (with $\sigma$$<$1\%).}
\label{fig:shuffle chart}
\end{figure}
Inspired by recent observations in~\citet{pham-2020-empirical}, we perturb the \texttt{numset} by shuffling the words of the corresponding question in each sample. We create three randomly permuted sets, that differ in the n-gram permutation.  1-gram shuffle refers to all words being shuffled, 2-gram and 3-gram refer to corresponding shuffles of ordered 2 and 3 grams in the question.
For the question mentioned in the introduction, an example of the possible shuffles for each of the \{3,2,1\}-grams are:
\vspace*{-2mm}
\begin{description}[leftmargin=0cm,style=sameline, noitemsep]
\item[3-gram] \textit{10000 about 1100 from did it take to decrease to How many years for the population? }
\item[2-gram] \textit{population to it take How many about 1100 from 10000 for the years did decrease to?}
\item[1-gram] \textit{10000 for from years to decrease it population about 1100 did many to take How the?}
\end{description}
\vspace*{-2mm}

As the random permutations may distort the semantic meaning (and destroy the syntax) of the question, we expect the predictions of the models to be severely impacted (with results approaching that of random chance). However, our experiments reveal (in Figure~\ref{fig:shuffle chart}) that this is not the case. Generally the trend suggests that 3-gram permutations do not generally degrade the performance severely. While we notice a that 1-gram permutation degrades the performance, we still observe that the models tend to predict a significant portion of questions correctly ($>$35\% F1 score). 
Some models are barely affected, such as NAQANet. We remark that most models that are based on BERT or bag-of-embeddings seem to be generally more robust to permutation than RoBERTa based models.

We present an analysis on  the effect of the number of numerical attributes in the passage and the ability of the models to make correct predictions in Table~\ref{table:perturbation-breakdown}. 
We bin questions into quartiles, such that each bin contains the same number of questions. The first bin contains passages with at most 12 numerical attributes, the second bin has between 13 and 18 numerical attributes, the third ranges between 19 and 23, and finally, the last bin contains passages with more than 23 numerical attributes.
\begin{table}[h]
\adjustbox{width=\columnwidth}{%
\begin{tabular}{@{}p{1.25cm}llp{1.25cm}p{1.25cm}p{1.25cm}@{}}
\toprule
                       &                                   & \textbf{Original} & \textbf{Shuffled 3-gram} & \textbf{Shuffled 2-gram} & \textbf{Shuffled 1-gram} \\ \midrule
\multirow{4}{*}{\textbf{MTMSN}} & $|n| \leq 12 $              & 79.27             & 75.4                     & 67.54                    & 50.05                    \\
                       & $12 < |n| \leq 18$ & 76.93             & 71.39                    & 65.39                    & 47.14                    \\
                       & $18 < |n| \leq 23$ & 81.23             & 76.06                    & 70.96                    & 54.88                    \\
                       &$|n|>23$            & 74.47             & 68.523                   & 63.76                    & 48.29                    \\\midrule 
\multirow{4}{*}{\textbf{NeRd}}  &  $|n| \leq 12 $               & 78.42             & 74.42                    & 70.21                    & 60.05                    \\
                       & $12 < |n| \leq 18$ & 78.85             & 74.16                    & 70.44                    & 61.37                    \\
                       & $18 < |n| \leq 23$ & 81.65             & 76.92                    & 74.14                    & 64.76                    \\
                       & $|n|>23$            & 77.11             & 71.77                    & 67.21                    & 58                       \\\midrule 
\multirow{4}{*}{\textbf{TASE}}  &  $|n| \leq 12 $               & 81.16             & 74.42                    & 62.82                    & 34.91                    \\
                       & $12 < |n| \leq 18$ & 77.86             & 70.25                    & 62.04                    & 34.17                    \\
                       &$18 < |n| \leq 23$ & 82.28             & 75.33                    & 67.55                    & 38.98                    \\
                       & $|n|>23$            & 75.39             & 66.89                    & 59.09                    & 29.26                    \\ \midrule 
\end{tabular}
}
\caption{Breakdown of models' performance (F1 score) on question-perturbation experiment based on the number of numerical attributes in the passage ($|n|$), compared to performance on original \texttt{numset}. }
\label{table:perturbation-breakdown}
\end{table}
We observe that there is no clear association between the performance of the models and the number of numerical attributes in a passage.
This experiment indicates that the models are not sensitive to word-order and this can potentially impact their utility.

\paragraph{Affinity to the class of questions}

We further probe models on their affinity to answer questions by only relying on the \emph{class} of questions. As the questions in DROP follow a certain pattern the type can potentially be inferred by exploiting the first few words. In this experiment, we only make available the first few words of the question to the models. This typically contains insufficient details and should make it difficult for the models to arrive at the \emph{correct} answer (average length of questions in \texttt{numset} is 12 words). 
We evaluate over three settings: passing the first two words, passing the first three words, and passing the first five words as the corresponding question. Below is an example of each on the question mentioned earlier: 
\vspace*{-2mm}
\begin{description}[leftmargin=0cm,style=sameline, noitemsep]
\item[2 Words] \textit{How many? }
\item[3 Words] \textit{How many years?}
\item[5 Words] \textit{How many years did it?}
\end{description}
\vspace*{-2mm}

\begin{figure}[ht]
\centering
\adjustbox{width=\columnwidth}{%
\includegraphics[width=\columnwidth]{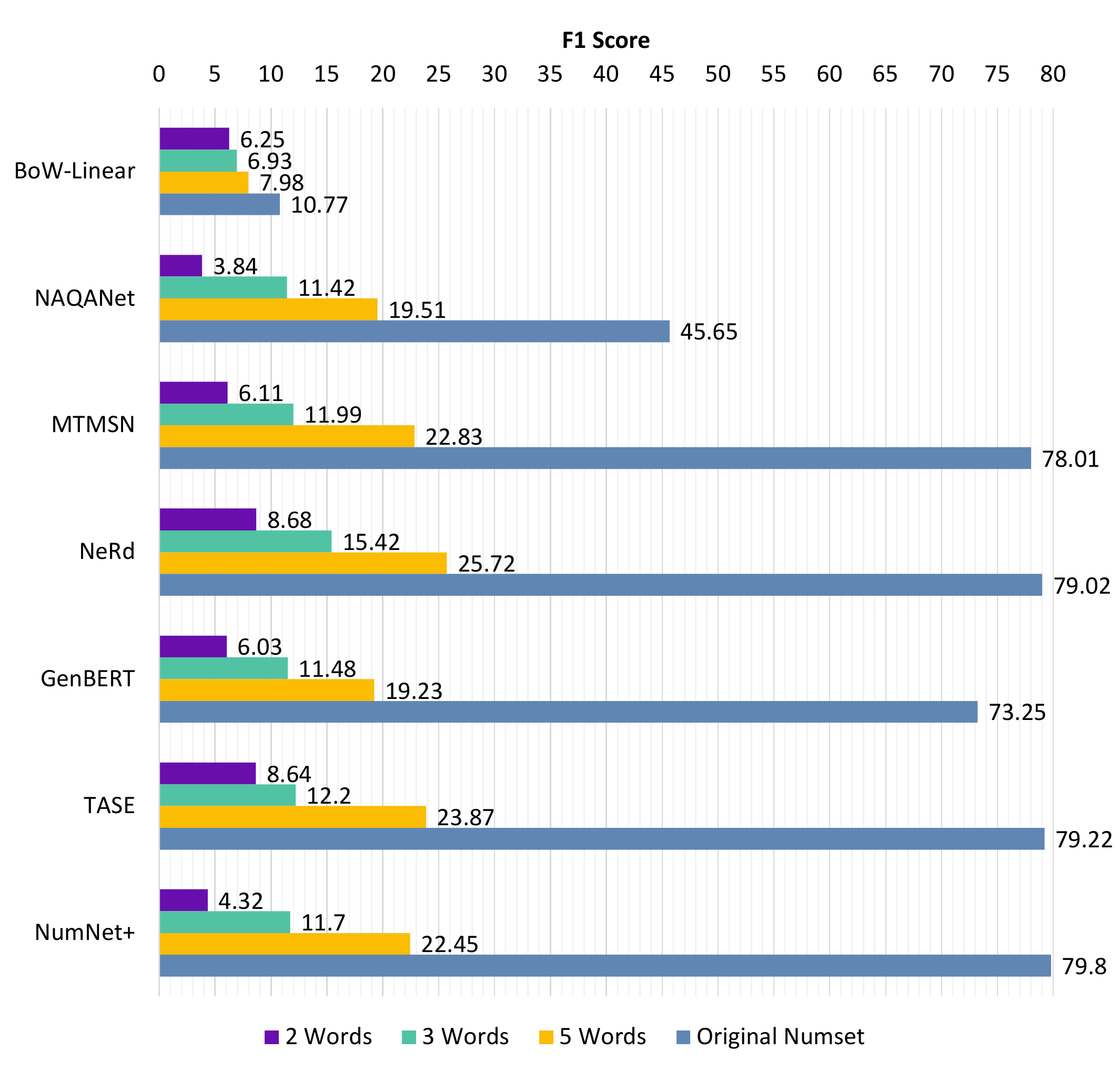}
}
\caption{F1 scores for each of the models on the three partial question settings and original. }
\label{fig:partialq chart}
\end{figure}

Fig~\ref{fig:partialq chart} shows the performance of the models on partial questions. With only five words, most models can still maintain a third of their correct predictions, and with only the first trigram of the question, they obtain an F1 $>$11.4\%, where NeRd obtains 15.42\%. Further showing an affinity of the models to be able to answer questions by exploiting the mere presence of a few words in the question.

 Unlike the permutation experiment, here breaking down performance based on existence of numerical attributes in the passage shows a steady decline in performance with more numbers in a passage (see Table~\ref{table:qaff-breakdown}), suggesting that it is more likely to get an answer correct if the space of possible arithmetic expressions is smaller.
\begin{table}[h]
\adjustbox{width=\columnwidth}{%
\begin{tabular}{@{}p{1.25cm}llp{1.25cm}p{1.25cm}p{1.25cm}@{}}
\toprule
                       &                                   & \textbf{Original} & \textbf{2Words} & \textbf{3Words} & \textbf{5Words} \\ \midrule
\multirow{4}{*}{\textbf{MTMSN}} 
                       & $|n| \leq 12 $     & 79.27 & 5.5 & 16.58 & 30.28 \\
                      & $12 < |n| \leq 18$ & 76.93          & 7.5          & 12.73          & 24.759         \\
                      & $18 < |n| \leq 23$ & 81.23          & 6.18         & 9.687          & 18.27          \\
                      & $|n|>23$           & 74.47          & 4.26         & 8.257          & 16.724         \\ \midrule
\multirow{4}{*}{\textbf{NeRd}} & $|n| \leq 12 $     & 78.42          & 10.33        & 22.59          & 32.79          \\
                      & $12 < |n| \leq 18$ & 78.85          & 9.62         & 14.77          & 28.11          \\
                      & $18 < |n| \leq 23$ & 81.65          & 8.4          & 12.65          & 21.57          \\
                      & $|n|>23$           & 77.11          & 5.93         & 10.99          & 19.01          \\ \midrule
\multirow{4}{*}{\textbf{TASE}} & $|n| \leq 12 $     & 81.16          & 11.29        & 17.27          & 29.11          \\
                      & $12 < |n| \leq 18$ & 77.86          & 8.08         & 11.79          & 25.915         \\
                      & $18 < |n| \leq 23$ & 82.28          & 8.94         & 10.94          & 21.948         \\
                      & $|n|>23$           & 75.39          & 5.97         & 8.25           & 17.354         \\ \midrule 
\end{tabular}
}
\caption{Breakdown of models' performance (F1 score) on affinity to class of questions experiment based on the number of numerical attributes in the passage ($|n|$), compared to performance on original \texttt{numset}. }
\label{table:qaff-breakdown}
\end{table}


\section{Evaluating passage comprehension}
\label{sec:passage}
\begin{figure}[t]
\centering
\adjustbox{width=\columnwidth}{%
\includegraphics[width=\columnwidth]{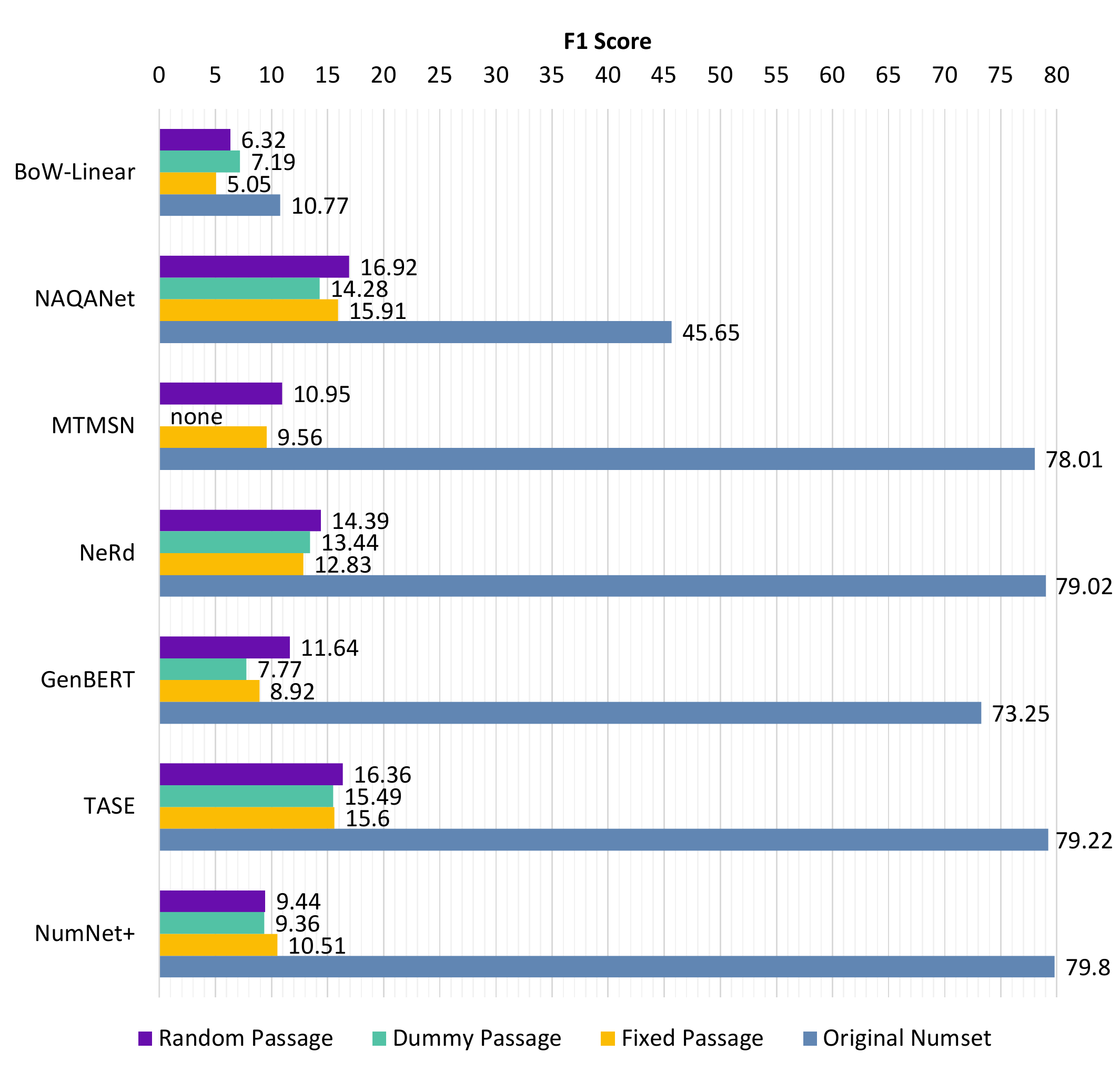}
}
\caption{F1 scores for each of the models on the three irrelevant passage settings and original. }
\label{fig:passage chart}
\end{figure}

We now examine whether the models are comprehending the passage and if they are basing their predictions on the evidence provided in the passage.
We probe with the following three settings:
\vspace*{-3mm}
\begin{description}[leftmargin=0cm,
style=sameline, noitemsep]
\item[Random Passage] We pair each question with a randomly assigned passage from \texttt{numset}.%
\item[Dummy Passage] We create an uninformative passage that contains no numbers, this is a proxy for a blank passage as models are unable to process that. It is the sequence: `\textit{This is a sentence.}' %
\item[Fixed Passage] We pair all questions with an unseen passage from the hidden test set. This passage has similar properties as passages in the train and dev, but is irrelevant to the corresponding question. %
\end{description}
\vspace*{-1mm}

Figure~\ref{fig:passage chart} shows the results of these three settings\footnote{Missing bars mean that the model failed to run.}; we observe a general trend where the models are able to correctly answer a significant portion of the questions without even having access to the relevant context (with F1 between 10\%---17\%). 
On further examination we observed that \texttt{span}-type questions cover the majority of the correctly predicted answers in this setting, indicating that \emph{comparison} type questions might carry inherent biases that models might exploit, such as questions of the form: 'which group is larger: households or people?', or even models picking up on the structure of the questions and learning to predict one of the two entities.
Table~\ref{table:unrelated-passage-comparison} shows the number of correctly predicted comparison questions in \texttt{numset} and the percentage of these that can still be predicted correctly if we take its context away. Most worryingly, we observe that NAQANet and NeRd maintain almost all of their predictions across the different settings and do not seem to take the textual context into account. NumNet+ is the only model whose performance degrades drastically. We hypothesize that this could be due to the GNN reasoning module (with each number in the passage appearing as a node) that informs its decision. GenBERT exhibits a curious behavior of a significant drop in performance for Dummy and Fixed Passage settings, while maintaining $>$56\% of its predictions in Random Passage. We postulate that this could be due to similarities in some passages in the devset.
\begin{table}[h]
\centering
\adjustbox{width=0.95\columnwidth}{%
\begin{tabular}{@{}lllll@{}}
\toprule
\textbf{Model} & \multicolumn{1}{{p{1.5cm}}}{\textbf{Original Dataset}} & \multicolumn{1}{p{1.5cm}}{\textbf{Random Passage}} & \multicolumn{1}{{p{1.5cm}}}{\textbf{Dummy Passage}} & \multicolumn{1}{{p{1.5cm}}}{\textbf{Fixed Passage}} \\ \midrule
NAQANet        & 635                                           & 97.80\%                                     & 95.43\%                                    & 94.96\%                                    \\
MTMSN          & 728                                           & 56.18\%                                     & \multicolumn{1}{l}{-}                      & 58.24\%                                    \\
NeRd           & 758                                           & 72.03\%                                     & 72.43\%                                    & 71.64\%                                    \\
GenBERT        & 704                                             &56.71\%                         & 24.93\%                     &   36.85\% \\
TASE           & 829                                           & 63.09\%                                     & 59.83\%                                    & 55.61\%                                    \\
NumNet+        & 807                                           & 22.18\%                                     & 16.48\%                                    & 27.51\%                                    \\ \bottomrule
\end{tabular}
}
\caption{Percentage of questions correctly predicted after replacing the passage with an unrelated one on comparison-type questions.}
\label{table:unrelated-passage-comparison}
\end{table}

\section{Related work}
\label{sec:rel-work}
Recent work has shown the effects of word-permutation on the performance of BERT-based models in NLU tasks \citep{pham-etal-2020-order, sinha-etal-2021-unnatural, sinha-etal-2021-masked, Gupta-etal-2021-bert}.  Chiefly, \citet{pham-etal-2020-order} analyzed the effect of word-order on 6 binary-classification GLUE tasks and demonstrated the limitations of BERT-based models. In our work, we investigate it in the context of numerical reasoning and observe similar behavior, but more generally in transformer-based models. 
Investigating NLP models in the context of NLU has been the focus of several recent works. We briefly highlight the prominent related works that include  \citet{jia-liang-2017-adversarial} who show that span-extraction RC models can be fooled by adding an adversarial sentence to the end of the passage,  
~\citet{mccoy-etal-2019-right} identify superficial heuristics that NLI models exploit instead of deeply understanding the language, and 
\citet{ ravichander-etal-2019-equate} evaluate their quantitative reasoning NLI benchmark on SoTA models and find that they are similar to a majority-class baseline. \citet{rozen-etal-2019-diversify} finds that a BERT-based NLI model fails to generalize to unseen number ranges in an adversarial dataset measuring numerical reasoning, suggesting an inherent model weakness. 

Contrast Sets \citep{gardner-etal-2020-evaluating} and Semantics Altering Modifications (SAMs) \citep{schlegel-etal-2020-sams} are two works that introduce changes to MRC benchmarks to better understand the decision boundaries of the models. They include a subset of the DROP dataset, wherein the former the benchmark's authors manually modify questions to include more compositional reasoning steps or change their semantics to create a Contrast Set. In the latter, the authors introduce an automatic way of generating SAMs, which alter the semantics of a sentence while preserving most of its lexical surface form. In our work, we take an inverse approach, where we alter the surface form of a question such that it no longer carries the meaning of the original question.

\section{Discussion and conclusion}
\label{sec:discussion}
In this work, we closely examined some of the top-performing models for numerical reasoning on DROP. 
Our study suggests that models are not necessarily arriving at the correct answer by reasoning about the question and content of the passage. Both question understanding and passage comprehension experiments reveal serious holes in the way the models are able to arrive at the correct answers. 
We hypothesize that the models have managed to pick up on the spurious patterns of the benchmark, rather than solving the task. Possible reasons for biases include: patterns in the format of passages and questions, where passages either describe the outcome of an American football match, the census of a certain location, or some historical event \citep{gardner-etal-2020-evaluating}, resulting in redundancies in the structure of a passage and patterns in their content; and answer frequency distribution, with top 5 answers being shared between train and dev splits, covering almost 20\% of the data. 
In fact, we found that in the \emph{affinity to the class of questions} experiments, there exists a vast disparity in performance between questions with most-frequently occurring answers vs. others. In NeRd, for example, the EM for questions with 2~words is 17.75\% in top-10 answers, whereas it is 4.2\% in the rest. The disparity narrows with more words, as it is 21.5\% vs. 11.90\%  for questions with 3~words, and 30.81\% vs. 22.23\% for questions with 5~words.

Benchmark leaderboards as they stand now can be misleading, incentivizing models to improve upon the reported scores without solving the underlying task. We strongly advocate for better methods to assess the capability of models for numerical reasoning. One such direction could be akin to \citet{linzen-2020-accelerate} who proposes a parallel evaluation paradigm that rewards models for possessing human-like generalization capabilities and \citet{Liu-etal-2021-explainaboard} that augments current leaderboards with three extra dimensions of interpretability, interactivity, and reliability. We highly recommend for  
careful design of the benchmarks and better leaderboards to correctly measure progress in such complex tasks. 





\section*{Acknowledgements}
This research has been supported by a PhD scholarship from King Saud University. We thank our anonymous reviewers for their constructive comments and suggestions, and SPIKE research group members for their feedback throughout this work.

\clearpage
\bibliography{anthology,emnlp}
\bibliographystyle{acl_natbib}
\appendix
\section{Random Baselines}
\label{appendix:random-baselines}
To ground the models' results, we evaluate two random baselines on the \texttt{numset}. The first randomly samples a pair of numbers from the passage, and finds their absolute difference (since subtraction is most prevalent) -- EM $1.8\%$. The second random baseline samples a final answer proportional to the frequency of answers in the training set -- EM $2.46\%$.
\section{Raw Results}
\label{sec:raw}

\begin{table}[h]
\adjustbox{width=\columnwidth}{%
\begin{tabular}{@{}lllllll@{}}
\toprule
                                & \multicolumn{2}{p{2cm}}{\textbf{Shuffled 3-gram}} & \multicolumn{2}{p{2cm}}{\textbf{Shuffled 2-gram}} & \multicolumn{2}{p{2cm}}{\textbf{Shuffled 1-gram}} \\ \midrule
\textbf{Model} & EM               & F1               & EM               & F1               & EM               & F1               \\ \midrule
BOW-Linear & 9.86           & 10.10          & 9.86           &10.11     &9.94   &10.21 \\
NAQANet                         & 39.45            & 41.86            & 36.30            & 39.14            & 34.18            & 37.17            \\
MTMSN                           & 71.95            & 72.90            & 65.85            & 66.91            & 48.50            & 50.00            \\
NeRd                            & 73.24            & 74.36            & 69.31            & 70.55            & 59.85            & 61.09            \\
GenBERT                         & 67.94            & 69.04            & 64.77            & 66.00            & 54.17            & 55.80            \\
TASE                            & 70.23            & 71.79            & 60.67            & 62.90            & 31.12            & 34.42            \\
NumNet+                         & 72.81            & 73.77            & 64.33            & 65.53            & 43.26            & 45.57            \\ \bottomrule
\end{tabular}
}
\caption{Models' performance when on each of the n-gram shuffles. }
\label{table:shuffles}
\end{table}

\begin{table}[h]
\adjustbox{width=\columnwidth}{%
\begin{tabular}{@{}lllllll@{}}
\toprule
        & \multicolumn{2}{l}{\textbf{2 Words }} & \multicolumn{2}{l}{\textbf{3 Words}} & \multicolumn{2}{l}{\textbf{5 Words}} \\ \midrule
\textbf{Model}   & EM                   & F1                   & EM                & F1               & EM                & F1               \\ \midrule
BOW-Linear & 6.06           &6.25          &6.76           &6.93     &7.90   &7.98 \\
NAQANet &  3.31     &3.84                & 10.9              & 11.42            & 18.97             & 19.51            \\
MTMSN   &  5.42     &6.11                 & 11.55             & 11.99            & 22.15             & 22.83            \\
NeRd    &  8.00     &8.68                & 14.94             & 15.42            & 25.08             & 25.72            \\
GenBERT &  5.43     &6.03                  & 10.88             & 11.48            & 18.54             & 19.23            \\
TASE    & 7.96                    & 8.64                    & 11.59             & 12.2             & 23.16             & 23.87            \\
NumNet+ &  3.68     &4.32                & 11.00             & 11.7             & 21.68             & 22.45            \\ \bottomrule
\end{tabular}
}
\caption{Models' performance when given only a few words from the beginning of the sentence. }
\label{table:some-words}

\end{table}

\begin{table}[h]
\adjustbox{width=\columnwidth}{%
\begin{tabular}{lllllll}
\toprule
                       & \multicolumn{2}{p{2cm}}{\textbf{Random Passage}} & \multicolumn{2}{p{2cm}}{\textbf{Dummy Passage}} & \multicolumn{2}{p{1.5cm}}{\textbf{Fixed Passage}} \\ \midrule
\textbf{Model} & EM                   & F1                   & EM                   & F1                  & EM              & F1              \\ \midrule
BOW-Linear & 6.22          &  6.32         & 6.45           &7.19     &4.69  &5.05 \\
NAQANet                         & 15.92                & 16.92                & 13.04                & 14.28               & 14.79           & 15.91           \\
MTMSN                           & 9.64                 & 10.95                & -                    & -                   & 8.10            & 9.56            \\
NeRd                            & 13.67                & 14.39                & 12.85                & 13.44               & 12.09           & 12.83           \\
GenBERT                         & 11.19                & 11.64                & 7.68                 & 7.77                & 8.64            & 8.92            \\
TASE                            & 15.23                & 16.36                & 14.59                & 15.49               & 14.41           & 15.6            \\
NumNet+                         & 8.46                 & 9.44                 & 8.57                 & 9.36                & 9.18            & 10.51           \\\bottomrule
\end{tabular}
}
\caption{Models' performance when given an unrelated passage, in three variations. }
\label{table:unrelated-passage}

\end{table}



\end{document}